\documentclass{article}

\usepackage[english]{babel}

\usepackage[letterpaper,top=2cm,bottom=2cm,left=3cm,right=3cm,marginparwidth=1.75cm]{geometry}

\usepackage{amsmath}
\usepackage{graphicx}
\usepackage{parskip}
\usepackage[colorlinks=true, allcolors=blue]
{hyperref}

\newcommand{\lora}{LoRA}
\newcommand{\loras}{LoRA }

\begin{document}

\title{A Note on \lora}

\author{
Vlad Fomenko\thanks{work done while at Microsoft}
\and Han Yu \and Jongho Lee \and Stanley Hsieh
\and Weizhu Chen\thanks{Weizhu completed the majority of the manuscript, Vlad edited the manuscript and drafted section 2.1} 
}
\date{Microsoft
\\
vlfomenk@gmail.com wzchen@microsoft.com
}

\maketitle

\begin{abstract}
\noindent \loras (Low-Rank Adaptation) \cite{hu2021lora} has emerged as a preferred method for efficiently adapting Large Language Models (LLMs) with remarkable simplicity and efficacy. This note extends the original \loras paper by offering new perspectives that were not initially discussed and presents a series of insights for deploying \loras at scale. Without introducing new experiments, we aim to improve the understanding and application of \lora.
\end{abstract}

\section{Additional Insights}

\subsection{On Comparison} 
Although the original \loras paper compared \loras with a variety of alternative methods,  it didn't fully explain why we designed \loras in such a way or how it tackles the challenges born in other approaches. 

Back in 2020, the predominant parameter-efficient adaptation technique was Adapter \cite{houlsby2019parameterefficient}. This method sequentially integrates two adaptation modules in each Transformer \cite{vaswani2017attention} layer, one after the attention and the other after the feed-forward modules.
This not only leads to extra inference latency, particularly with smaller batch sizes as highlighted in the \loras study, but it also causes a significant increase in the network's depth.
Empirically, we observed that this increase often led to training instability. Specifically, for certain tasks or datasets, achieving training convergence became challenging, particularly when working with the 96-layer GPT-3 model \cite{brown2020language}. The issue with increased depth partly inspired us to consider expanding the network in its width rather than its depth, which laid the foundation for \lora's design that extends weights in parallel, contrasting with the Adapter's sequential approach.

Around the same time, a separate project led by Yang and Hu et al \cite{hpt} on hyper-parameter transfer (HPT), demonstrated the practicality of transferring hyperparameters across a model's width. However, attempts to apply HPT along the model's depth were less successful.
This lent further credence to the rationale behind extending networks in parallel width, as \loras does, rather than sequentially like the Adapter. Indeed, there was a lack of comprehensive evidence or theory explaining the difficulties in either model adaptation or hyper-parameters transfer in terms of depth. This gap in understanding is a reason why we initially refrained from discussing such perspectives in the original \loras paper. 

During our exploration of \lora, we concurrently examined Prefix Tuning \cite{li2021prefixtuning} and Prompt Tuning \cite{lester2021power}. Although Prefix Tuning offered a novel approach, its reduction of the model's context length posed a significant limitation. In contrast, Prompt Tuning, despite showing potential, delivered inconsistent outcomes across different datasets in our tests. This underscored that input-level modifications may not suffice for ensuring stability and consistency in diverse applications and that changes in the model's internal structure are crucial.

\loras distinguishes itself by implementing adaptations at the matrix level, a more streamlined approach compared to the Adapter's addition of extra layers.
This granular level of adaptation allows \loras to be versatile and applicable to various modules, including different matrices within Transformers' attention layers, the fully connected layers in the Feed-Forward Network (FFN) blocks, and even the embedding layers. This makes \loras broadly applicable to any model relying on matrix computations.

\subsection{On Motivation}
One of the main initial motivations for exploring efficient fine-tuning from an infrastructure standpoint was a considerable network burden due to the costs of transferring model weights and optimization states, especially over cross-regional networks. Such issues often arise during the saving and loading of checkpoints. While caching the weights of a static pretrained model can mitigate the need to re-download weights for fine-tuning, supporting continual fine-tuning or resuming a pre-existing paused experiment necessitates frequent re-fetching of model weights. Moreover, this challenge is exacerbated for large-scale models that require a distributed training setup across multiple nodes. This also increases the risk of network failure during weight transfer. Consider training a GPT-3 model equipped with 175 billion parameters and FP16 weights. Its snapshot occupies approximately 350GB, necessitating the use of multiple nodes to manage the weights and their optimizer states, either in RAM or via networked storage. Checkpointing the weights of such a distributed model can introduce a lot of overhead. Yet, transitioning to \loras significantly stabilizes checkpoint management during training, as it only requires saving and transferring the comparatively smaller \loras matrices. For continual fine-tuning, with \loras employed, it is no longer required to download the entire model weights, but just the relevant \loras matrices, assuming the base model weights were pre-existing or cached beforehand (e.g., from a previous run). 
While initially, we believed that enhancing the training stability was the primary benefit, we soon discovered that deploying \loras models at scale for online inference yielded even more significant and relevant advantages. We will explain this in more detail in a subsequent section.  

\subsection{On FFN}
The original \loras paper puts a primary focus on the attention layers, with a limited examination of its effects on the Feed-forward Network (FFN) module in Transformers \cite{vaswani2017attention}. Initially, we encountered inconsistencies in FFN performance using \lora, leading to a reduced interest in further FFN investigations. However, several months after publishing the original paper, we identified and rectified a bug in our \loras FFN implementation. Subsequent extensive experimentation revealed that applying \loras to FFN can be effective and often complements attention-based \lora. Nonetheless, considering the additional memory demands of \lora, attention-based \loras typically offers greater efficacy within memory constraints. We provide more insights on the placement of \loras in Transformers below.

\section{Practical Improvements}

Below we discuss insights and practices learned over the past several years of extensive deployment of models trained with \loras in production.

\subsection{Placement}
\label{placement}

\lora's versatility enables it to be applied across a variety of model architectures that perform matrix multiplication operations. Our insights primarily derive from applying \loras within Transformers for NLP tasks, where the choice of placement can significantly influence training outcomes.

The optimal placement for \loras is highly dependent on the dataset and model architecture, with the size of the model being a critical factor. While uniformly applying \loras to all matrices yields the best training outcomes in most cases, we often achieved comparable performance by selectively applying \loras to a subset of matrices. The optimal selection varied across tasks and architectures. For some datasets, especially those of a larger scale, the performance gap between \loras and full fine-tuning could not be fully bridged. This suggests the necessity for customized experiments tailored to each unique scenario.

In our experience, applying \loras exclusively to attention layers provides the most stability and mitigates the risk of divergence, albeit at the cost of requiring multiple training epochs for optimal performance. The next effective target for \loras application has been the embedding matrices, especially for smaller-scale models where these matrices constitute a larger proportion of parameters. When \loras was applied to un-embedding matrix, the addition of \loras to the embedding matrix often became redundant. Incorporating \loras into the fully connected (MLP) layers can further enhance model performance. As for hyperparameters, we observed that the default values generally performed well for \loras training, however, when \loras was applied to a small subset of matrices, higher values for learning rate were required. Overall, adjustment of \loras placement can maintain the balance between the model's capacity, speed of adaptation, and the risk of overfitting.

Investigating \loras applied to MoE (Mixture of Experts) models, we found that applying \loras to each expert individually boosted performance in many setups. Yet, this approach significantly increased memory usage, making it less cost-effective. We observed limited success with applying \loras to the router matrix, which only benefited certain setups.

The effectiveness of \loras is also influenced by the base model's size. As the model scale increases, the benefits of using a larger \loras rank saturate faster, and the performance gap between the most effective \loras setup and full fine-tuning diminishes. This suggests a strategy of applying \loras to as many matrix types as feasible before considering increasing \loras rank, within memory constraints. Further memory optimization can be achieved by leveraging techniques such as sharing the same $B$ matrix across different $A$ matrices in \lora, e.g., for the attention matrices $W_Q$, $W_K$, and $W_V$ in Transformers.

In summary, there is no one-size-fits-all strategy for \loras placement. Our experience advocates for a progressive approach: starting with attention matrices, then embedding matrices, followed by fully-connected (MLP) matrices, and finally applying \loras across all matrices, while increasing its rank, until the desired performance is achieved. This approach balances the trade-offs between model quality, training time, and memory consumption during inference.

\subsection{Inference}
Previous studies often credit \loras for its efficiency in enhancing the training process. However, as we applied \loras in production at scale, we realized that a more significant impact stems from \lora's cost-effective online serving. Most notably, by serving \loras models with non-merged weights, one can reduce the cost of serving an additional \loras model to a minimal extent.

In general, there are three main ways to serve trained \loras models for inference. The first is to merge the \loras weights with the base weight to produce a checkpoint of the same format as the base model. This approach can offer zero extra inference latency, compared to serving the base model, since no extra operations are needed during inference. However, we rarely adopt this approach in production, unless the use case is extremely sensitive to inference latency and the same model needs to be deployed on a large number of GPUs, so that the fungibility of sharing GPUs across different \loras models is not crucial. Otherwise, this approach has several disadvantages. First, it introduces a large network overhead when transferring the full model weights for deployment. Second, it creates a deployment-time architecture mismatch, as during training, the model employed a separate pathway for \loras weights, prior to merging. It can also introduce numerical instability, especially when working with low-precision formats like 4-bit \cite{dettmers2023qlora}, since merging of the weights is lossy and non-trivial, e.g., often requiring re-quantization.

A straightforward alternative is to serve the resultant \loras model in a non-merged form, with the delta \loras weights explicitly present in the inference graph. This approach enables a single base model to dynamically pair with multiple delta \loras weights, i.e., multiple models. As the base model's weights remain intact, the same GPUs can keep them in memory, only swapping the \loras parts of the computational graph or loading multiple \loras weights at once, and masking out all but the currently selected weights. For every new request that requires a different \loras model, this approach allows for a fast weights swap operation to serve the new \loras model. Nevertheless, while the \loras delta weights are small, swapping them can still introduce a noticeable overhead for online serving, impacting latency, throughput, and serving costs.

The third option is to serve multiple models, i.e., \loras weights, on the same set of GPUs over a shared endpoint, routing incoming requests to the correct underlying delta \loras weights. Such a design can enable production services to serve thousands or even hundreds of thousands of \loras models, with the same base model, at once. Implementations of this design can also allow for a batch of requests to point to different \loras weights, which can be dynamically selected during the forward pass. Further optimization techniques, such as buffering and batching the incoming requests, can bring significant speedups. Since most inference operations are still memory-bound, batching multiple requests together is the key to better utilizing the GPU resources, significantly reducing the cost and increasing the overall throughput.

Below we describe one approach to enable serving multiple \loras models at once, which can support requests pointing to multiple \loras models, without swapping the \loras weights, while maintaining latency comparable to a request pointing to a single model. We first combine \loras weights for every shared base layer, from all \loras models, into a series of stacked tensors, one per each base layer. When treating a batch request pointing to multiple \loras models, we define a batched routing mask with weights of 1 assigned to the indices of the target \loras models' weights, from the stacked \loras matrices, while nullifying the rest. We implemented a set of kernels that support batched multiplication of such masks with stacked \loras weights, allowing for efficient forward passes with little overhead. This approach is reminiscent of the routing and Mixture-of-Expert (MoE) for the FFN layers in Switch Transformer \cite{fedus2022switch} and can enable efficient batch serving of requests targeting a large number of \loras models at once. Such system helped us to serve \loras at a production scale by reducing the additional latency and cost of a new \loras model to a minimal extent. A recent work, S-LoRA \cite{slora}, proposes a similarly effective solution for this scenario with several optimizations.

\subsection{Additional Explorations}
We have also investigated multiple methodologies beyond our primary focus, yet these explorations did not culminate in impactful outcomes.

A notable investigation involved the implementation of an adaptive version of \lora, where the rank dimension $r$ is dynamically determined for each layer and matrix during training. While this approach often helped to enhance the model's quality, it was constrained by increased training duration and infrastructure challenges during inference. Specifically, such an approach resulted in a higher levels of memory fragmentation, causing larger overheads during inference. Batching \loras requests for models with varying \loras dimensionality posed a further problem. The recent development of S-\loras \cite{slora} may offer a solution to these challenges, suggesting the potential for future adoption of Adaptive \lora.

We also explored augmenting the vanilla \loras with various techniques, such as non-linearity \cite{he2022unified}, similar to the DenseNet \cite{huang2018densely}, but for \loras weights only, expanding \loras into MoE \loras \cite{moelora}, or combining \loras with other parameter-efficient training techniques \cite{mao2022unipelt}. 

While some approaches improved the results on certain datasets, their increased complexity hindered the ease of integrating \loras with base models. When the model size was large enough, our observations indicated that non-linearity added to \loras did not substantially benefit performance, and MoE \loras was not sufficiently cost-effective due to the additional memory requirements.

As outlined in our original publication, attempts were made to combine \loras with other techniques, like Prefix Tuning and Prompt Tuning, given their orthogonal nature in structural augmentation. However, we ultimately favored the simplicity and maintainability of using \loras exclusively, considering its ease of future extensions and exploring the application of \loras to different matrices at once, as detailed in section 2.1.

\section{Looking Ahead}
Despite its popularity and various advantages, there are many opportunities to make \loras and other parameter-efficient fine-tuning methods even more effective for both research and production.

First, when a model, on which \loras weights were based, is changed or updated, the current methodology would require re-training all the \loras models, diminishing the method's utility. Finding a viable solution for this issue remains elusive, complicating the upkeep of services that utilize numerous \loras models, when base models need to be updated monthly or annually.

Second, although \loras often outperforms other methods during inference, it remains relatively slow and expensive in training, particularly for large-scale models. Preliminary attempts to create \loras parameters without backpropagation \cite{hypertuning} show potential but are not effective enough for practical use yet. Other studies \cite{lorahub} \cite{shah2023Zip\lora} explored developing new \loras models from pre-existing \loras weights, instead of starting from scratch. Further innovation in \loras synthesis is necessary to enhance quality and adaptability for varied tasks in a production environment.

The rise of quantization-aware training introduces new complexities. While low-precision training with \loras \cite{dettmers2023qlora} represents a significant advancement in enabling \loras to run on low-memory GPUs, it also quantizes the model weights, which can degrade the performance. Recent studies \cite{loftq} \cite{guo2023lqlora} attempt to bridge this gap by integrating the quantization discrepancy into \lora's initial weights. These results are preliminary, and further research is essential, especially as quantized training is poised to gain widespread popularity.

Although \loras originated from a study of language modeling tasks, it has been successfully applied to models and tasks for other modalities, especially for computer vision tasks, e.g., to diffusion models \cite{rombach2022highresolution}. Further research on combining the simplicity and effectiveness of \loras with the distinct mechanisms inherent to such methods, e.g., the multi-step denoising in diffusion models, is likely to yield exciting advancements.

\subsection*{Acknowledgments}
We would like to thank Edward Hu for proofreading the draft and providing edit suggestions. 

\bibliographystyle{alpha}
\bibliography{main}

\end{document}